\documentclass[11pt]{article}

\usepackage[preprint]{acl}
\usepackage{array}
\usepackage{booktabs}
\usepackage{multirow}
\usepackage{graphicx}
\graphicspath{{figures/}}
\usepackage[table]{xcolor}
\usepackage{pifont}
\usepackage{siunitx}
\usepackage{listings}
\usepackage{mdframed}
\usepackage{bbm}
\usepackage{amsmath}
\newcommand{\method}{\texttt{GraphRAG-LTM}}
\usepackage[most]{tcolorbox}
\usepackage{stfloats}
\usepackage{algorithm}
\usepackage{algorithmic}
\usepackage{float}
\usepackage{needspace}

\newcommand{\contribding}[3]{%
  \par\addvspace{0.2em}%
  \noindent
  \hangindent=1.2em
  \hangafter=1
  \ding{#1}\ \textbf{\textit{#2}} #3\par%
}

\newtcblisting{prompt}{
  listing only,             % CRITICAL: Printing only the code box, not running it
  breakable,                % CRITICAL: Allows the box to split across pages
  before skip=10pt,         % Space above the prompt block
  after skip=10pt,          % Space below the prompt block
  enhanced,                 % Advanced drawing engine (required for breakable)
  colback=gray!5!white,     % Light gray background
  colframe=gray!50!black,   % Dark gray border
  boxrule=0.4pt,            % Border thickness
  arc=2pt,                  % Corner radius
  boxsep=2pt,               % Internal margin
  left=2pt, right=2pt, top=2pt, bottom=2pt, % Padding
  listing options={
    basicstyle=\ttfamily\small, % Monospaced, small font
    breaklines=true,            % Wrap text at end of line
    breakatwhitespace=false,    % Allow splitting words if necessary
    columns=fullflexible,       % Natural character spacing
    mathescape=false,           % Treat $ as a character, not math
    literate={-}{-}1,           % Proper rendering of dashes
  }
}

\usepackage{xparse}
\usepackage{xfp} % for \fpeval
\ExplSyntaxOn
\NewDocumentCommand{\dcell}{mm}{ % #1=value, #2=base
  \fp_set:Nn \l_tmpa_fp {#1 - #2}
  \fp_set:Nn \l_tmpb_fp {abs(\l_tmpa_fp)}
  \num{#1}%
  \raisebox{-0.75ex}{\hspace{0.15em}\scriptsize
    \fp_compare:nNnTF {\l_tmpa_fp} = {0}
      {\textcolor{gray!70}{$\rightarrow$\num{\fp_eval:n{round(\l_tmpb_fp,1)}}}}
      {
        \fp_compare:nNnTF {\l_tmpa_fp} < {0}
          {\textcolor{red!70!black}{$\downarrow$\num{\fp_eval:n{round(\l_tmpb_fp,1)}}}}
          {\textcolor{green!50!black}{$\uparrow$\num{\fp_eval:n{round(\l_tmpb_fp,1)}}}}
      }
  }
}
\ExplSyntaxOff

\usepackage[table]{xcolor}   % rowcolor / cellcolor
\usepackage{booktabs}        % \toprule \midrule \bottomrule / \specialrule
\usepackage{siunitx}         % if you use S columns elsewhere (optional here)

\definecolor{tableheadcolor}{HTML}{F2E6E6}

\definecolor{blue1}{HTML}{D2E6FA}
\colorlet{blue2}{blue1!50}

\newcommand{\topcell}[1]{\cellcolor{blue1}\textbf{#1}}
\newcommand{\secondcell}[1]{\cellcolor{blue2}#1}

\usepackage[dvipsnames]{xcolor}

\newcommand{\downchg}[1]{%
  \kern0.08em\raisebox{-0.65ex}{\scriptsize\textcolor{BlueGreen}{\ensuremath{\downarrow}#1}}%
}

\newcommand{\upchg}[1]{\kern0.08em\raisebox{-0.65ex}{\scriptsize\textcolor{RedOrange}{\ensuremath{\uparrow}#1}}}

\definecolor{blue1}{HTML}{d2e6fa}
\colorlet{blue2}{blue1!50}
\definecolor{blue1}{HTML}{d2e6fa}
\colorlet{blue2}{blue1!50}

\definecolor{headerbg}{HTML}{F2E6E6}
\definecolor{modelbg}{HTML}{D5D4D6}

\usepackage{times}
\usepackage{latexsym}

\usepackage[T1]{fontenc}

\usepackage[utf8]{inputenc}

\usepackage{microtype}

\usepackage{inconsolata}

\usepackage{graphicx}

\title{GraphRAG-LTM: A Layered Threat Modeling Framework for Attacks on Graph Retrieval-Augmented Generation}

\author{
\textbf{Yining Tang}\textsuperscript{1} \quad \textbf{Junyu Zhang}\textsuperscript{1} \quad
\textbf{Haomiao Qiu}\textsuperscript{2,3} \quad \textbf{Yongchao Liu}\textsuperscript{4} \\
\textbf{Jiajun Zheng}\textsuperscript{4} \quad \textbf{Zhao Wang}\textsuperscript{1} \quad
\textbf{Ziyue Qiao}\textsuperscript{3,$\dagger$} \quad \textbf{Chao Wu}\textsuperscript{1,$\dagger$} \\
\textnormal{\small \textsuperscript{1}Zhejiang University \quad
\textsuperscript{2}Harbin Institute of Technology, Shenzhen} \\
\textnormal{\small \textsuperscript{3}Great Bay University \quad \textsuperscript{4}Ant Group} \\
\textnormal{\small \textsuperscript{$\dagger$}Corresponding authors: \texttt{zyqiao@gbu.edu.cn}, \texttt{chao.wu@zju.edu.cn}}
}

\begin{document}
\maketitle
\begin{abstract}
Graph retrieval-augmented generation (GraphRAG) uses document-derived knowledge graphs to assemble evidence for large language model (LLM) answers.
Its security depends on both the seed chunks retrieved for a query and the graph evidence assembled from them, yet attacks on this evidence path remain underexplored.
We introduce \method{}, a \underline{L}ayered \underline{T}hreat \underline{M}odeling framework that tests three nested levels of attacker access.
BB Query uses ranking feedback to make query edits under a semantic-similarity constraint and demote relevant seed chunks.
WB Query uses retriever gradients for the same objective.
Graph-aware WB keeps the white-box adversarial query and deletes bridge triplets from the query-specific evidence graph, disrupting multi-hop evidence propagation through localized graph edits.
Across datasets, retrievers, and LLM backbones, rank feedback alone reduces answer quality, gradient access strengthens query attacks, and bridge-triplet deletion further impairs supporting-fact retrieval.
Further analyses show how seed retrieval and graph evidence jointly contribute to these failures.
\end{abstract}

\section{Introduction}

\begin{figure}[t]
    \centering
    \includegraphics[width=1\linewidth]{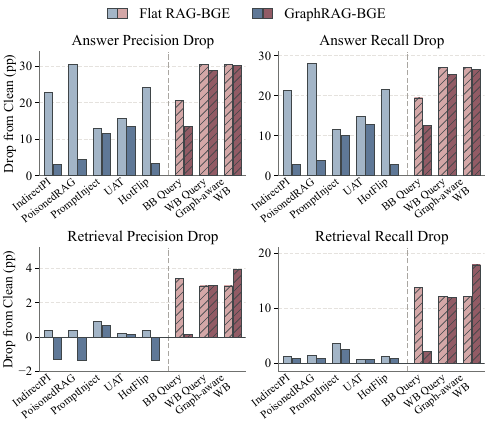}
\caption{
Answer and supporting-fact retrieval precision and recall drops (percentage points from Clean) on HotpotQA-Distractor with BGE and LLaMA3-8B. Blue bars show prior baselines; red hatched bars show our BB Query, WB Query, and Graph-aware WB attacks. Lighter and darker shades denote Flat RAG and GraphRAG, respectively. Several baselines cause substantially larger drops in answer precision and recall on Flat RAG than on GraphRAG, whereas our attacks remain effective on GraphRAG.
}
    \label{fig:intro_teaser}
\end{figure}

Retrieval-augmented generation (RAG) grounds large language model (LLM) answers in external evidence~\cite{lewis2020retrieval,gao2024retrieval,fan2024survey}. Recent \emph{GraphRAG} methods further use document-derived knowledge graphs to organize inter-chunk dependencies and support multi-hop context construction~\cite{wang2024knowledge,edge2024local,zhu2025knowledge}. In these pipelines, retrieved seed chunks determine which pre-extracted triplets enter a query-specific evidence graph and, ultimately, which evidence reaches the generator.

Existing attacks have studied corpus poisoning, malicious document injection, and indirect prompt injection~\cite{zhong2023poisoning,cao20245w1h,greshake2023not}, mostly in \emph{Flat RAG} with unstructured retrieval. In GraphRAG, a shifted seed ranking can also alter graph expansion and the final context, allowing retrieval errors to propagate through evidence selection. Figure~\ref{fig:intro_teaser} reports our experiments applying prior attacks and our proposed perturbations to Flat RAG and GraphRAG on the same questions. Several prior attacks produce large drops in answer precision and recall in Flat RAG but substantially smaller drops in GraphRAG, whereas perturbations to seed retrieval and graph evidence continue to degrade GraphRAG.

\begin{figure}[H]
    \centering
    \noindent\includegraphics[width=\columnwidth]{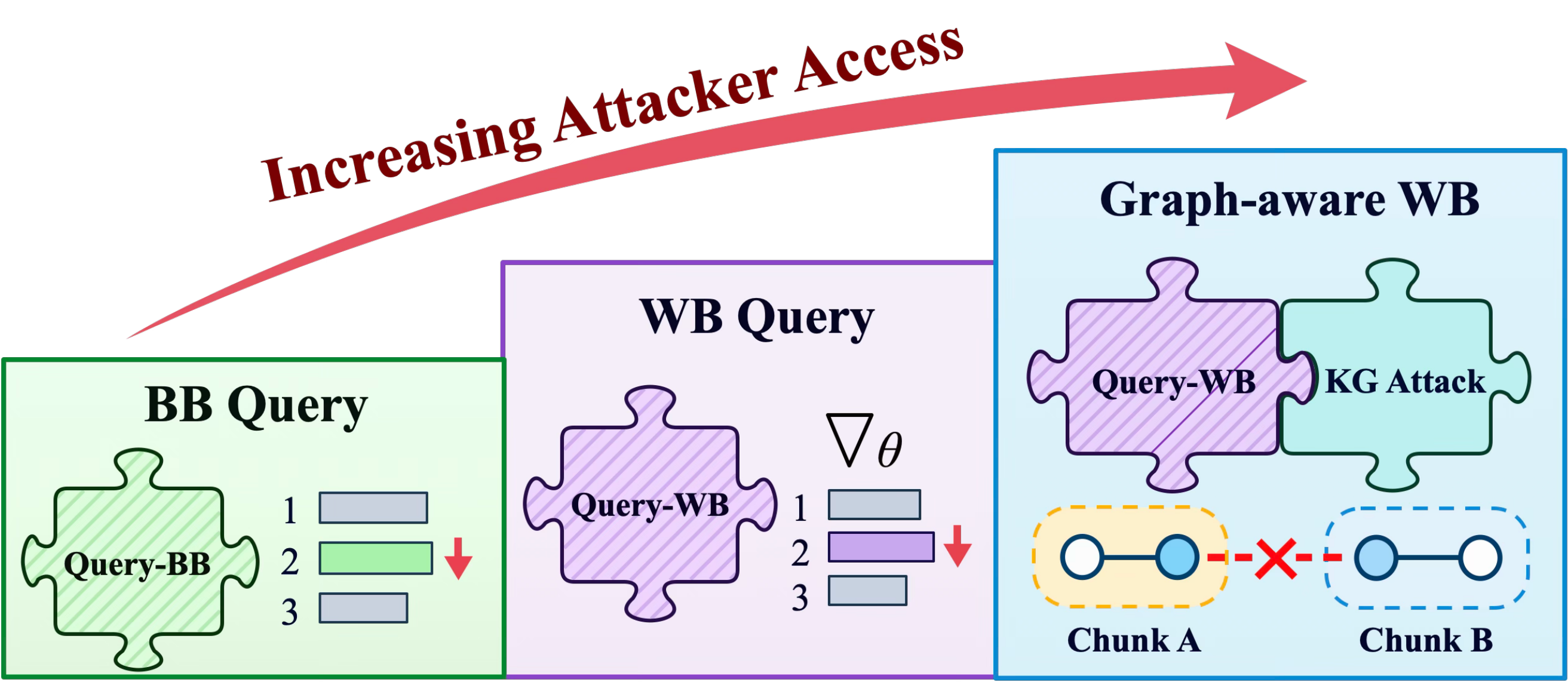}\par
    \vspace{0.3em}
    \small
    \setlength{\tabcolsep}{4pt}
    \renewcommand{\arraystretch}{1.1}
    \noindent\resizebox{\columnwidth}{!}{%
    \begin{tabular}{@{}lccc@{}}
        \toprule
        \textbf{Access} & \textbf{BB Query} & \textbf{WB Query} & \textbf{Graph-aware WB} \\
        \midrule
        Rank feedback & \ding{51} & \ding{51} & \ding{51} \\
        Retriever gradient & \ding{55} & \ding{51} & \ding{51} \\
        KG triplets & \ding{55} & \ding{55} & \ding{51} \\
        \bottomrule
    \end{tabular}%
    }
    \caption{Three nested attacker access levels in \method{}, from rank feedback to query-specific graph editing. The diagram shows attack components; the table lists each tier's permissions.}
    \label{fig:threat-access-tiers}
\end{figure}

These observations raise the question of how adversarial perturbations at different stages of GraphRAG's evidence pipeline affect the evidence it retrieves and the answers it generates. Figure~\ref{fig:threat-access-tiers} frames this question through three nested attacker access levels, labeled \emph{BB Query}, \emph{WB Query}, and \emph{Graph-aware WB}. At the first level, the attacker observes rank feedback; at the second, the attacker additionally accesses retriever gradients; at the third, the attacker can further inspect and edit KG triplets.

We introduce \method{} to instantiate these access levels with corresponding attacks. At the first two levels, a ranking-guided query attack uses semantically constrained substitutions to demote originally retrieved seed chunks, relying on rank feedback or retriever gradients according to the available access. At the third level, the attack reuses the white-box adversarial query and deletes bridge triplets in the query-specific evidence graph. This matched design measures the added effect of graph edits beyond query perturbation. Our evaluation across multiple datasets, retrievers, LLM backbones, and graph-retrieval mechanisms reveals a layered failure pattern: rank feedback alone can degrade answer generation, retriever gradients strengthen query attacks, and bridge-triplet deletion further impairs supporting-fact retrieval. Our contributions are summarized as follows:
\contribding{182}{Layered Threat Modeling for GraphRAG.}{The contrasting effects of several existing attacks in Flat RAG and GraphRAG motivate a threat model centered on GraphRAG's evidence pipeline. We formulate three nested access levels and evaluate a corresponding attack at each level to distinguish what rank feedback, retriever gradients, and graph-triplet access each enable.}

\contribding{183}{Ranking-guided Query Attack on Graph-construction Anchors.}{
Seed chunks anchor the query-specific evidence graph and downstream context selection. We develop a ranking-guided query attack that demotes them through semantically constrained edits. It uses ranking feedback in the black-box setting and retriever gradients in the white-box setting.
}

\contribding{184}{Graph-aware Extension via Bridge Triplets.}{
To examine what graph access adds to the query attack, we delete bridge triplets while keeping the white-box adversarial query fixed across the paired conditions. This comparison isolates the additional effect of graph edits on evidence retrieval, which we further examine under two alternative graph-retrieval mechanisms that differ in how they select and organize evidence.
}

\section{Related Work}
\noindent\textbf{LLM safety.}
Hallucination and factuality remain central LLM reliability concerns~\cite{huang2025survey,yehuda2024interrogatellm,kryscinski-etal-2020-evaluating}, motivating the use of retrieved evidence to ground generation.
Other work studies misuse and failures across the model lifecycle~\cite{wang2025comprehensive,mcguffie2020radicalization,ma2023adapting}.
Guardrails, rule-guided feedback, and self-correction address model behavior~\cite{rebedea2023nemo,mu2024rule,gou2023critic}; retrieval-based systems also depend on the reliability of external evidence supplied at inference time.

\noindent\textbf{Retrieval-augmented generation.}
RAG augments LLMs with retrieved documents to ground answers and access external knowledge across a broad range of tasks~\cite{lewis2020retrieval,guu2020retrieval,izacard2023atlas}.
Noisy or misranked passages can displace relevant evidence and harm generation~\cite{karpukhin2020dense,izacard2021unsupervised,cuconasu2024power}.
Query refinement, ranking-feedback rewriting, and active retrieval aim to improve evidence selection~\cite{chan2024rq,mao2024rafe,jiang2023active}.
GraphRAG systems organize text and graph evidence for multi-hop reasoning or query-focused summarization~\cite{edge2024local,jin2024graph,mavromatis2024gnn,zhu2025knowledge}, using different mechanisms to select and compose evidence.

\noindent\textbf{Adversarial attacks on LLMs and RAG.}
Prompt injection places adversarial instructions in text read by the generator~\cite{liu2023prompt,greshake2023not}.
Corpus poisoning inserts malicious passages, while prompt perturbations and coordinated prompt--corpus attacks provide other ways to steer generated answers~\cite{zou2402poisonedrag,hu2024prompt,jiao2025pr}.
At the text level, synonym substitutions use semantic-similarity constraints~\cite{ren2019pwws,jin2020textfooler}, while HotFlip uses gradients to identify discrete edits~\cite{ebrahimi2018hotflip}.
Query-level RAG studies also show that small query variations can affect retrieval and downstream QA~\cite{percin2025queryrobustness}.
We examine how constrained query edits shift seed rankings and how graph-triplet edits further change evidence retrieval under stronger attacker access.

\begin{figure*}[t]
  \centering
  \includegraphics[width=\textwidth]{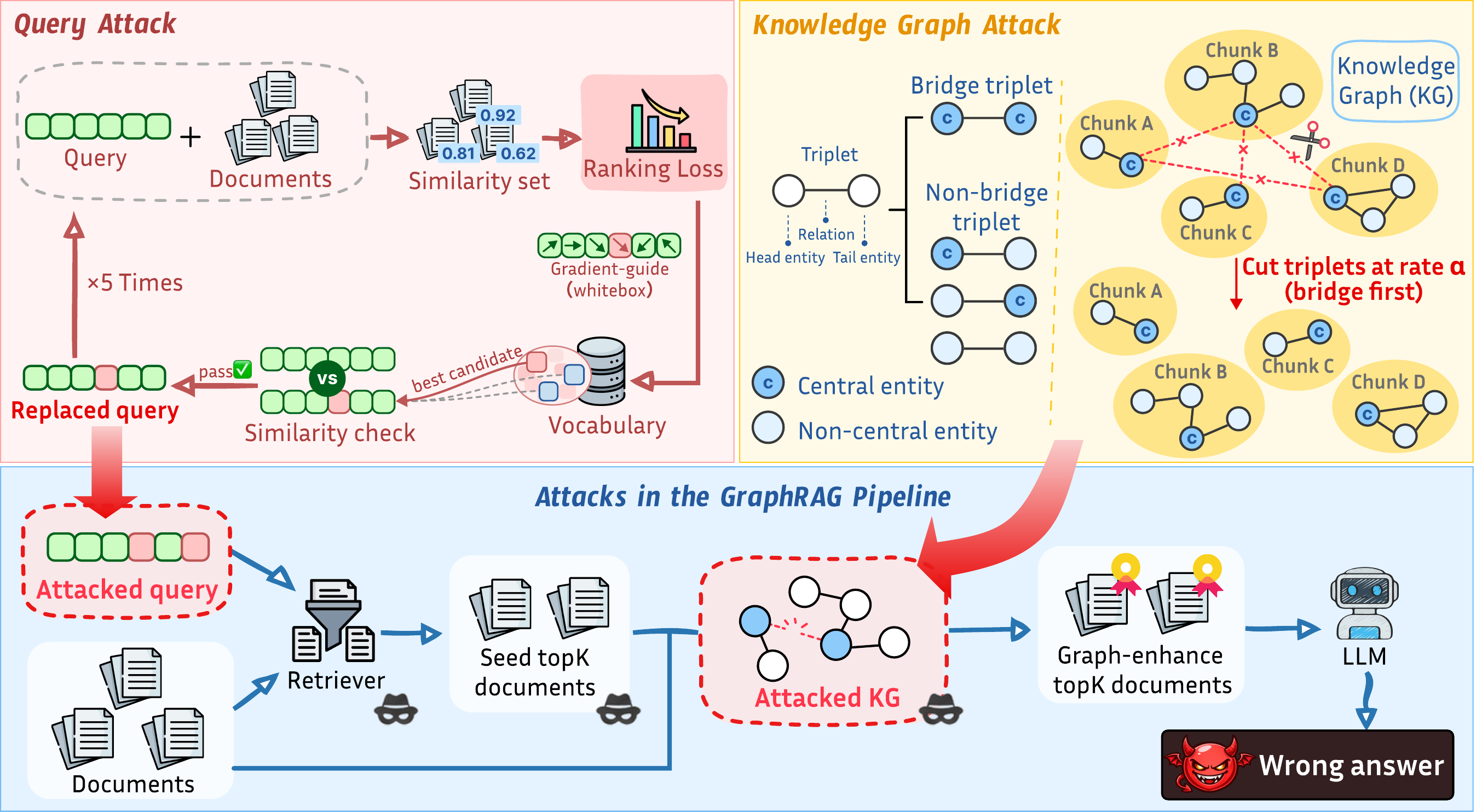}
  \caption{
  Attack components across the three access tiers of \method{}.
  The upper-left panel shows the query attacks of Tiers 1 and 2: BB Query uses ranking feedback, whereas WB Query uses the gradient-guided edits shown; both aim to demote seed chunks.
  Tier 3 retains WB Query and adds the bridge-triplet cuts shown in the upper-right panel.
  The bottom panel maps these interventions onto the GraphRAG pipeline: query edits affect seed retrieval at all three tiers, while bridge-triplet cuts affect graph-based evidence expansion only at Tier 3.
  }
  \label{fig:total-attack}
\end{figure*}

\section{Problem Settings and Threat Models}
\label{sec:problem}

\subsection{GraphRAG Pipeline}
\label{subsec:graphrag-pipeline}

We study GraphRAG pipelines that organize retrieved evidence with document-derived knowledge triplets.
Given text chunks $\mathcal{D}=\{c_i\}_{i=1}^{n}$, the system extracts triplets $(h,r,t)$ from each chunk and forms a knowledge graph $\mathcal{T}$, where triplets remain linked to their source chunks.
For a query $Q$, a semantic retriever returns the top-$K$ seed chunks from the document collection, $\mathcal{S}_Q = \pi_K(Q;\mathcal{D})$, where $\pi_K(\cdot)$ denotes the retriever ranking.

The system then constructs a seed subgraph by collecting triplets associated with $\mathcal{S}_Q$ and expands it through graph traversal or reranking to obtain a query-specific evidence graph $\mathcal{G}_Q$.
A context organizer selects the final context $C_Q$ from the retrieved chunks and graph-expanded evidence, and the generator $\mathcal{M}$ produces $A = \mathcal{M}(Q, C_Q)$.
The attack objective is to degrade the final answer quality, measured by the discrepancy between $A$ and the gold answer.
Unlike Flat RAG, the seed chunks in GraphRAG not only provide initial textual evidence but also determine the downstream evidence graph and final context.
Therefore, studying attacks along the query-to-graph pathway is critical for understanding the robustness of GraphRAG systems. 

\iffalse
\subsection{Attack Objective}
\label{subsec:attack-objective}

The attacker aims to degrade the final answer while keeping perturbations constrained.
Let $Q'$ denote an adversarial query, and let $\mathcal{T}'$ denote a perturbed graph when graph-aware access is available.
The general attack objective is:
\begin{align}
    &\max_{Q',\mathcal{T}'}
    \mathcal{L}_{\mathrm{attack}}(A', A^\star) \label{eq:attack-objective}\\
    &\quad \mathrm{s.t.} \quad
    \mathrm{sim}(Q,Q') \geq \theta_{\mathrm{sim}}, 
    \quad
    d(\mathcal{T},\mathcal{T}') \leq \delta_{\mathrm{graph}} . \notag
\end{align}
Here $A^\star$ is the gold answer, $\mathrm{sim}(Q,Q')$ constrains semantic drift, and $d(\mathcal{T},\mathcal{T}')$ constrains graph edits.
Different threat models allow different variables to be modified: query-only attacks keep $\mathcal{T}'=\mathcal{T}$, while graph-aware attacks may additionally perturb triplets in the query-specific evidence graph.
\fi

\subsection{Layered Threat Models}
\label{subsec:threat-models}

We consider three progressively stronger attacker access levels at different stages of the GraphRAG pipeline, as illustrated in Figure~\ref{fig:threat-access-tiers}.

\paragraph{Black-box Ranking-feedback Query Attack.}
The attacker observes retrieved rankings or top-$K$ results and perturbs the seed retrieval anchor without accessing retriever internals or graph triplets.
This setting models an external user of a GraphRAG service that exposes ranked source passages or retrieval results.

\paragraph{White-box Gradient-guided Query Attack.}
The attacker uses retriever gradients to guide query edits but cannot inspect or modify graph triplets.
This setting applies when retriever parameters are available to the attacker, as in an open-source system or a compromised retrieval service.

\paragraph{Graph-aware White-box Propagation Attack.}
Building on WB Query, the attacker can inspect and cut bridge triplets in the query-specific evidence graph.
This setting models an insider or compromised graph-processing component with access to per-query graph artifacts, without control of the full corpus or global graph.

\section{The Proposed \method{}}
\label{sec:method}
\subsection{Overview}
\label{subsec:method-overview}

\method{} demotes originally relevant seed chunks through constrained query edits, so ranking shifts can propagate into graph expansion and final context selection.
Its \emph{BB Query} and \emph{WB Query} variants share a retrieval-anchor objective but use rank feedback and retriever gradients, respectively; \emph{Graph-aware WB} adds bridge-triplet perturbation to the same white-box query.
Figure~\ref{fig:total-attack} shows where these interventions enter the GraphRAG pipeline.

\subsection{Retrieval-anchor Ranking Objective}
\label{subsec:ranking-query}

The query-side attack targets the graph-construction anchor of GraphRAG.
Let $\mathcal{I}_{\mathrm{orig}}=\pi_K(Q)$ denote the top-$K$ seed chunks retrieved by the clean query $Q$.
Our goal is to construct an adversarial query $Q'$ such that chunks in $\mathcal{I}_{\mathrm{orig}}$ are demoted under the adversarial ranking $\pi_K(Q')$, while limiting semantic drift, $\operatorname{sim}(Q,Q') \ge \theta_{\mathrm{sim}}$.
Similar minimum-similarity constraints are used in adversarial text editing, including TextFooler~\cite{jin2020textfooler}.
This objective is shared by \emph{BB Query}, \emph{WB Query}, and the query-side component of \emph{Graph-aware WB}.

Only the top-$K$ retrieved chunks seed graph expansion.
We therefore factor the loss contribution of each original seed $d_i\in\mathcal{I}_{\mathrm{orig}}$ into three terms:
\ding{172} a clean-rank importance weight $w_i$;
\ding{173} a prominence score $\phi_i$ measuring how highly it remains ranked under $Q'$;
and \ding{174} a selection mask $\omega_i$ emphasizing whether it remains in the top-$K$ results.
The ranking loss is:
\begin{equation}
\mathcal{L}_{\mathrm{rank}}(Q')
=
\frac{1}{K}
\sum_{d_i\in\mathcal{I}_{\mathrm{orig}}}
w_i\cdot \phi_i \cdot \omega_i .
\label{eq:ranking-loss}
\end{equation}
Minimizing this loss suppresses the original seed evidence, thereby shifting both retrieval and downstream graph construction.

\paragraph{Importance Weight.}
Both \emph{BB Query} and \emph{WB Query} use the same importance weight $w_i$, which depends only on the clean-query ranking and gives larger weights to higher-ranked chunks.
Let $r_i(Q)$ denote the clean rank of $d_i$.
We use a DCG-inspired weighting scheme:
\begin{equation}
w_i=
\begin{cases}
\displaystyle
\frac{2^{K-r_i(Q)+1}-1}
{\sum_{j=1}^{K}(2^{K-j+1}-1)}
& \text{if } r_i(Q)\le K,\\[6pt]
0 & \text{otherwise}.
\end{cases}
\label{eq:dcg-weight}
\end{equation}
This prioritizes suppressing highly ranked clean evidence, since these chunks are more likely to anchor the evidence graph.

\subsection{{\fontfamily{pcr}\selectfont BB Query}: Rank-only Black-box  Attack}
\label{subsec:blackbox-query}

At the first access tier, \emph{BB Query} instantiates Eq.~\ref{eq:ranking-loss} with a discrete proxy computed from observed ranks.
Let $r_i(Q')$ be the observed rank of target chunk $d_i$ under the adversarial query.
If $d_i$ does not appear in the retrieved list, its contribution is set to zero for that candidate query.
The black-box prominence and mask are defined as:
\begin{align}
&\phi_i^{\mathrm{bb}}=\max(K-r_i(Q')+1,0),
\\
&\omega_i^{\mathrm{bb}}=\mathbf{1}[r_i(Q')\le K].
\end{align}
The black-box ranking loss is:
\begin{equation}
\mathcal{L}_{\mathrm{rank}}^{\mathrm{bb}}(Q')
=
\frac{1}{K}
\sum_{d_i\in\mathcal{I}_{\mathrm{orig}}}
w_i\cdot \phi_i^{\mathrm{bb}}\cdot \omega_i^{\mathrm{bb}}.
\label{eq:bb-loss}
\end{equation}

Starting from $Q'=Q$, we iteratively generate word-level substitution candidates from a synonym set, such as WordNet.
For each candidate query $\widetilde{Q}$, we query the retriever, compute $\mathcal{L}_{\mathrm{rank}}^{\mathrm{bb}}(\widetilde{Q})$, and accept the substitution only if $\operatorname{sim}(Q,\widetilde{Q})\ge \theta_{\mathrm{sim}}$ and $\mathcal{L}_{\mathrm{rank}}^{\mathrm{bb}}(\widetilde{Q})<\mathcal{L}_{\mathrm{rank}}^{\mathrm{bb}}(Q')$.
The process stops after reaching the substitution budget $T_{\max}$ or when no valid edit further reduces the discrete loss.
This gives a rank-only attack that perturbs the seed retrieval anchor without accessing model internals.

\subsection{{\fontfamily{pcr}\selectfont WB Query}: Retriever White-box  Attack}
\label{subsec:whitebox-query}
At the second access tier, \emph{WB Query} optimizes the retrieval-anchor objective with gradients from the target retriever.
Since exact ranks are non-differentiable, we first approximate the rank of each target chunk using pairwise comparisons.
Let $\mathcal{I}_{\mathrm{adv}}=\pi_{K+M}(Q')$ be the candidate pool retrieved by the adversarial query, where $M$ provides additional candidates beyond the top-$K$.
For each target chunk $d_i$, let $s_i=\langle q',d_i\rangle$ be its similarity score with the adversarial query representation $q'$.
We approximate its adversarial rank as:
\begin{equation}
\hat{r}_i(Q')
=
1+
\sum_{j\in\mathcal{I}_{\mathrm{adv}}, j\ne i}
\sigma\left(\frac{s_j-s_i}{\tau}\right),
\label{eq:soft-rank}
\end{equation}
where $\tau$ is a temperature parameter.
We then define the differentiable prominence and mask as:
\begin{align}
&\phi_i=\max(K-\hat{r}_i(Q')+1,0),
\\
&\omega_i=\sigma\left(\frac{K-\hat{r}_i(Q')}{\epsilon}\right).
\end{align}
Substituting these terms into Eq.~\ref{eq:ranking-loss} yields a differentiable ranking loss for white-box optimization.

Starting from $Q'=Q$, we compute the gradient of the ranking loss for each token position $k$ as $\mathbf{g}_k=\nabla_{\mathbf{e}_k}\mathcal{L}_{\mathrm{rank}}(Q')$, where $\mathbf{e}_k$ is the embedding of the current token.
For each candidate replacement token $v\in\mathcal{V}$ with embedding $\mathbf{e}_v$, we estimate the first-order effect of substitution by:
\begin{equation}
s_{k,v}
=
-(\mathbf{e}_v-\mathbf{e}_k)^\top \mathbf{g}_k.
\label{eq:grad-score}
\end{equation}
A larger $s_{k,v}$ indicates a stronger expected reduction of the ranking loss.

At each step, we select the candidate pair $(k,v)$ with the largest score and construct a candidate query $\widetilde{Q}$.
The substitution is accepted only if $\operatorname{sim}(Q,\widetilde{Q})\ge \theta_{\mathrm{sim}}$ and $\mathcal{L}^{wb}_{\mathrm{rank}}(\widetilde{Q})<\mathcal{L}^{wb}_{\mathrm{rank}}(Q')$.
The optimization stops after $T_{\max}$ successful substitutions or when no valid candidate further reduces the loss.
The white-box query optimization procedure is summarized in Algorithm~\ref{alg:whitebox}.

\begin{algorithm}[t]
\small
\caption{WB Query Perturbation}
\label{alg:whitebox}
\begin{algorithmic}[1]
\REQUIRE Query $Q$, retriever $\mathcal{R}$, max edits $T_{\max}$, similarity threshold $\theta_{\mathrm{sim}}$, vocabulary $\mathcal{V}$
\ENSURE Adversarial query $Q'$
\STATE Initialize $Q'\gets Q$, $t\gets 0$, modified positions $\mathcal{M}\gets\emptyset$
\WHILE{$t<T_{\max}$ and $|\mathcal{M}|<|Q|$}
    \FOR{each position $k\notin\mathcal{M}$}
        \STATE Compute $\mathbf{g}_k=\nabla_{\mathbf{e}_k}\mathcal{L}_{\mathrm{rank}}(Q')$
        \FOR{each candidate token $v\in\mathcal{V}$}
            \STATE Compute $s_{k,v}=-(\mathbf{e}_v-\mathbf{e}_k)^\top\mathbf{g}_k$
        \ENDFOR
    \ENDFOR
    \STATE Select $(k^*,v^*)=\arg\max_{k,v}s_{k,v}$
    \STATE Construct $\widetilde{Q}$ by replacing token $k^*$ with $v^*$
    \IF{$\operatorname{sim}(Q,\widetilde{Q})\ge\theta_{\mathrm{sim}}$ and $\mathcal{L}_{\mathrm{rank}}(\widetilde{Q})<\mathcal{L}_{\mathrm{rank}}(Q')$}
        \STATE $Q'\gets\widetilde{Q}$; $t\gets t+1$
        \STATE $\mathcal{M}\gets\mathcal{M}\cup\{k^*\}$
    \ELSE
        \STATE Mark $k^*$ as failed; add it to $\mathcal{M}$ if retry limit is reached
    \ENDIF
\ENDWHILE
\RETURN $Q'$
\end{algorithmic}
\end{algorithm}

\subsection{{\fontfamily{pcr}\selectfont Graph-aware WB}: Bridge-Triplet Cuts}
\label{subsec:graph-aware}
At the third access tier, the attacker can access the query-specific evidence graph and combine \emph{WB Query} with bridge-triplet cuts to disrupt multi-hop evidence propagation across retrieved chunks.

GraphRAG often relies on cross-chunk triplets to connect evidence from chunks.
We distinguish such connectors from local intra-chunk relations.
Let $\mathcal{E}_{\mathrm{docs}}$ denote document-level central entities.
A triplet $T_i=(h,r,t)$ is defined as a \emph{bridge triplet} with respect to a central entity $\mathtt{ent}$ if it connects $\mathtt{ent}$ to another document-level entity:
\begin{align}
\mathrm{is\_bridge}(T_i,\mathtt{ent})
&=
\big(\mathtt{ent}\in\{h,t\}\big)
\land
\label{eq:is-bridge}
\\
&\quad
\big(\{h,t\}\setminus\{\mathtt{ent}\}\subseteq \mathcal{E}_{\mathrm{docs}}\big).
\notag
\end{align}
Bridge triplets serve as connectors for multi-hop evidence propagation across chunks.

Given the query-specific subgraph $\mathcal{T}_{\mathcal{S}}$ and a deletion ratio $\alpha$, we first identify the bridge-triplet set $\mathcal{I}_{\mathrm{bridge}}$ according to Eq.~\ref{eq:is-bridge}.
We set the deletion budget to $N_{\mathrm{delete}}=\lfloor\alpha|\mathcal{T}_{\mathcal{S}}|\rfloor$.
We prioritize sampling from $\mathcal{I}_{\mathrm{bridge}}$ to target cross-chunk evidence connections.
If the number of bridge triplets is insufficient, we sample additional non-bridge triplets from $\mathcal{T}_{\mathcal{S}}\setminus\mathcal{I}_{\mathrm{bridge}}$ to meet the budget.
The selected triplets are cut to obtain the perturbed subgraph $\mathcal{T}'_{\mathcal{S}}$.

\section{Experiments}
\label{sec:experiments}

In this section, we conduct experiments under different attacker access levels to evaluate the adversarial robustness of GraphRAG systems. 
Our study focuses on whether perturbing the retrieval anchor and graph structure can effectively degrade graph-enhanced retrieval and downstream answer generation. 
Specifically, our experiments are designed to answer the following research questions:
\textbf{(RQ1)} How does attacker access level, from rank-only feedback to retriever gradients and graph-aware access, affect the vulnerability of GraphRAG?
\textbf{(RQ2)} How does bridge-triplet perturbation affect graph-enhanced retrieval and evidence propagation?
\textbf{(RQ3)} When does the attack take effect, and does its impact persist across perturbation budgets, graph deletion ratios, and generator backbones?
\textbf{(RQ4)} Does the graph-side effect persist under other graph-based evidence retrieval mechanisms?

\begin{table*}[t]
\centering
\newcommand{\tableonechg}[1]{\kern0.08em\raisebox{-0.55ex}{\footnotesize\textcolor{BlueGreen}{\ensuremath{\downarrow}#1}}}
\newcommand{\tableonebest}[1]{\kern0.08em\raisebox{-0.55ex}{\footnotesize\textcolor{BlueGreen}{\ensuremath{\downarrow}\textbf{#1}}}}
\setlength{\tabcolsep}{2.7pt}
\renewcommand{\arraystretch}{1.05}
\resizebox{\textwidth}{!}{%
\begin{tabular}{l|ccc|ccc|ccc|c}
\specialrule{1.2pt}{0pt}{0pt}
\rowcolor{headerbg}
& \multicolumn{3}{c|}{HotpotQA-Distractor} & \multicolumn{3}{c|}{MuSiQue} & \multicolumn{3}{c|}{2WikiMultihopQA} & \multicolumn{1}{c}{Overall} \\
\cline{2-4}\cline{5-7}\cline{8-10}\cline{11-11}
\rowcolor{headerbg}
\multirow{-2}{*}{\textbf{Method}} & Contriever & BGE & mxbai & Contriever & BGE & mxbai & Contriever & BGE & mxbai & Avg. \\
\specialrule{1.2pt}{0pt}{0pt}
\rowcolor{modelbg}
\multicolumn{11}{l}{\textbf{LLaMA3-8B}} \\
\rowcolor{gray!10}
Clean & 52.3 & 55.8 & 55.2 & 23.9 & 25.6 & 24.8 & 13.3 & 12.6 & 13.4 & 30.8 \\
PromptInject & 42.7\tableonechg{9.6} & 44.8\tableonechg{11.0} & 45.3\tableonechg{9.9} & 19.1\tableonechg{4.8} & 19.8\tableonechg{5.8} & 20.2\tableonechg{4.6} & 6.6\tableonechg{6.7} & 6.7\tableonechg{5.9} & 7.5\tableonechg{5.9} & 23.6\tableonechg{7.2} \\
\rowcolor{gray!10}
UAT & 39.2\tableonechg{13.1} & 42.6\tableonechg{13.2} & 41.4\tableonechg{13.8} & 14.0\tableonechg{9.9} & 15.1\tableonechg{10.5} & 15.3\tableonechg{9.5} & 6.0\tableonechg{7.3} & 6.9\tableonechg{5.7} & 7.0\tableonechg{6.4} & 20.8\tableonechg{10.0} \\
IndirectPI & 49.4\tableonechg{2.9} & 53.0\tableonechg{2.8} & 52.2\tableonechg{3.0} & 16.5\tableonechg{7.4} & 17.6\tableonechg{8.0} & 17.2\tableonechg{7.6} & 11.7\tableonechg{1.6} & 11.7\tableonechg{0.9} & 12.2\tableonechg{1.2} & 26.8\tableonechg{4.0} \\
\rowcolor{gray!10}
PoisonedRAG & 49.0\tableonechg{3.3} & 51.6\tableonechg{4.2} & 51.6\tableonechg{3.6} & 13.3\tableonechg{10.6} & 14.4\tableonechg{11.2} & 13.7\tableonechg{11.1} & 12.5\tableonechg{0.8} & 12.6 & 12.6\tableonechg{0.8} & 25.7\tableonechg{5.1} \\
HotFlip & 49.4\tableonechg{2.9} & 52.6\tableonechg{3.2} & 52.5\tableonechg{2.7} & 17.0\tableonechg{6.9} & 18.4\tableonechg{7.2} & 17.8\tableonechg{7.0} & 12.5\tableonechg{0.8} & 12.1\tableonechg{0.5} & 12.6\tableonechg{0.8} & 27.2\tableonechg{3.6} \\
RandKD & 52.4\upchg{0.1} & 55.3\tableonechg{0.5} & 55.1\tableonechg{0.1} & 23.9 & 25.6 & 24.8 & 13.4\upchg{0.1} & 13.1\upchg{0.5} & 13.7\upchg{0.3} & 30.8 \\
\specialrule{0.8pt}{0pt}{0pt}
\rowcolor{gray!10}
BB Query & 37.6\tableonechg{14.7} & 42.8\tableonechg{13.0} & 39.2\tableonechg{16.0} & 11.6\tableonechg{12.3} & 11.9\tableonechg{13.7} & 13.0\tableonechg{11.8} & 6.6\tableonechg{6.7} & 8.0\tableonechg{4.6} & 7.5\tableonechg{5.9} & 19.8\tableonechg{11.0} \\
WB Query & \secondcell{26.2}\tableonebest{26.1} & \secondcell{28.6}\tableonebest{27.2} & \secondcell{32.3}\tableonebest{22.9} & \secondcell{7.6}\tableonebest{16.3} & \secondcell{9.4}\tableonebest{16.2} & \secondcell{11.7}\tableonebest{13.1} & \secondcell{5.1}\tableonebest{8.2} & \topcell{5.1}\tableonebest{7.5} & \secondcell{6.0}\tableonebest{7.4} & \secondcell{14.7}\tableonebest{16.1} \\
\rowcolor{gray!10}
Graph-aware WB & \topcell{23.7}\tableonebest{28.6} & \topcell{27.4}\tableonebest{28.4} & \topcell{30.5}\tableonebest{24.7} & \topcell{7.4}\tableonebest{16.5} & \topcell{9.2}\tableonebest{16.4} & \topcell{11.4}\tableonebest{13.4} & \topcell{4.0}\tableonebest{9.3} & \secondcell{5.5}\tableonebest{7.1} & \topcell{5.9}\tableonebest{7.5} & \topcell{13.9}\tableonebest{16.9} \\
\specialrule{1.2pt}{0pt}{0pt}
\rowcolor{modelbg}
\multicolumn{11}{l}{\textbf{Qwen3-8B-Base}} \\
\rowcolor{gray!10}
Clean & 49.1 & 52.7 & 52.6 & 33.2 & 35.6 & 35.5 & 21.6 & 23.8 & 23.8 & 36.4 \\
PromptInject & 27.5\tableonechg{21.6} & 30.3\tableonechg{22.4} & 30.1\tableonechg{22.5} & 18.6\tableonechg{14.6} & 19.7\tableonechg{15.9} & 20.5\tableonechg{15.0} & 5.9\tableonechg{15.7} & \topcell{6.9}\tableonebest{16.9} & \topcell{6.6}\tableonebest{17.2} & 18.5\tableonechg{17.9} \\
\rowcolor{gray!10}
UAT & 42.7\tableonechg{6.4} & 46.6\tableonechg{6.1} & 46.6\tableonechg{6.0} & 23.8\tableonechg{9.4} & 26.2\tableonechg{9.4} & 26.1\tableonechg{9.4} & 16.2\tableonechg{5.4} & 18.4\tableonechg{5.4} & 18.5\tableonechg{5.3} & 29.5\tableonechg{6.9} \\
IndirectPI & 43.6\tableonechg{5.5} & 48.3\tableonechg{4.4} & 47.7\tableonechg{4.9} & 24.4\tableonechg{8.8} & 24.8\tableonechg{10.8} & 25.6\tableonechg{9.9} & 17.4\tableonechg{4.2} & 22.0\tableonechg{1.8} & 21.1\tableonechg{2.7} & 30.5\tableonechg{5.9} \\
\rowcolor{gray!10}
PoisonedRAG & 41.5\tableonechg{7.6} & 46.5\tableonechg{6.2} & 46.2\tableonechg{6.4} & 22.0\tableonechg{11.2} & 23.3\tableonechg{12.3} & 22.3\tableonechg{13.2} & 18.5\tableonechg{3.1} & 21.6\tableonechg{2.2} & 22.4\tableonechg{1.4} & 29.4\tableonechg{7.0} \\
HotFlip & 42.6\tableonechg{6.5} & 46.8\tableonechg{5.9} & 47.2\tableonechg{5.4} & 21.2\tableonechg{12.0} & 26.0\tableonechg{9.6} & 25.3\tableonechg{10.2} & 18.4\tableonechg{3.2} & 22.4\tableonechg{1.4} & 21.5\tableonechg{2.3} & 30.1\tableonechg{6.3} \\
\rowcolor{gray!10}
RandKD & 48.9\tableonechg{0.2} & 52.6\tableonechg{0.1} & 52.6 & 33.2 & 35.6 & 35.5 & 22.6\upchg{1.0} & 25.3\upchg{1.5} & 25.4\upchg{1.6} & 36.9\upchg{0.5} \\
\specialrule{0.8pt}{0pt}{0pt}
\rowcolor{gray!10}
BB Query & 33.2\tableonechg{15.9} & 33.4\tableonechg{19.3} & 34.2\tableonechg{18.4} & 15.1\tableonechg{18.1} & 15.2\tableonechg{20.4} & 15.9\tableonechg{19.6} & 10.4\tableonechg{11.2} & 9.9\tableonechg{13.9} & 10.0\tableonechg{13.8} & 19.7\tableonechg{16.7} \\
WB Query & \secondcell{23.7}\tableonebest{25.4} & \secondcell{26.6}\tableonebest{26.1} & \secondcell{27.8}\tableonebest{24.8} & \secondcell{9.8}\tableonebest{23.4} & \secondcell{12.5}\tableonebest{23.1} & \secondcell{13.4}\tableonebest{22.1} & \topcell{5.3}\tableonebest{16.3} & 8.4\tableonechg{15.4} & 7.9\tableonechg{15.9} & \secondcell{15.0}\tableonebest{21.4} \\
\rowcolor{gray!10}
Graph-aware WB & \topcell{20.9}\tableonebest{28.2} & \topcell{25.4}\tableonebest{27.3} & \topcell{26.7}\tableonebest{25.9} & \topcell{9.2}\tableonebest{24.0} & \topcell{12.1}\tableonebest{23.5} & \topcell{13.1}\tableonebest{22.4} & \secondcell{5.6}\tableonebest{16.0} & \secondcell{7.2}\tableonebest{16.6} & \secondcell{7.0}\tableonebest{16.8} & \topcell{14.1}\tableonebest{22.3} \\
\specialrule{1.2pt}{0pt}{0pt}
\end{tabular}}
\caption{Answer F1 (\%) across three datasets and three retrievers under two LLM settings. Arrows show percentage-point changes from the corresponding \textit{Clean} score. Within each LLM setting, dark/light blue highlight the lowest/second-lowest attacked F1 in each column, and the associated two declines are bold. Avg. is the unweighted mean of the nine dataset--retriever scores. Corresponding Retrieval F1 results are in Appendix~\ref{appendix:retrieval-f1}.}
\label{tab:answer-f1-extended}
\end{table*}

\subsection{Experiment Setup}
\label{subsec:exp-setup}

\paragraph{Datasets and Metrics.}
Our main evaluation uses 7,405 questions from HotpotQA-Distractor~\cite{yang2018hotpotqa}, 2,417 from MuSiQue~\cite{trivedi2022musique}, and a fixed 1,000-question development subset from 2WikiMultihopQA~\cite{ho2020constructing}, with graphs built from document text for all three datasets.
We report Answer F1 and Supporting-fact F1 as primary metrics; dataset preprocessing and evaluation details appear in Appendix~\ref{appendix:setup}.

\paragraph{GraphRAG Setup.}
For the main experiments, we instantiate GraphRAG with KG2RAG~\cite{zhu2025knowledge}.
This pipeline extracts document-derived triplets, retrieves the top-$K{=}10$ seed chunks, expands evidence through the graph, and selects the top 10 context chunks for generation.
We evaluate three retrievers: Contriever~\cite{izacard2021unsupervised}, BGE-large-en-v1.5~\cite{xiao2024cpack}, and mxbai-embed-large.
The main tables use two separate triplet-extraction and answer-generation settings, LLaMA3-8B~\cite{grattafiori2024llama} and Qwen3-8B-Base~\cite{yang2025qwen3}; other ablation studies and analyses use LLaMA3-8B unless otherwise stated.
Both model settings use the same extraction and generation prompt templates, reproduced in Appendix~\ref{appendix:prompts}.
Further implementation details and retrieval settings are provided in Appendix~\ref{appendix:setup}.

\paragraph{Attacks and Baselines.}
All attacks are applied at inference time.
For all main-result tables, we set the query-attack iteration budget to $T_{\max}{=}5$ under semantic similarity threshold $\theta_{\mathrm{sim}}{=}0.8$.
For graph-aware evaluation, Bridge Edge Cut removes $\alpha{=}5\%$ of triplets from the query-specific evidence graph.
We compare with Clean, PoisonedRAG~\cite{zou2402poisonedrag}, HotFlip corpus poisoning~\cite{ebrahimi2018hotflip,zhong2023poisoning}, Indirect Prompt Injection~\cite{greshake2023not}, PromptInject~\cite{perez2022ignore}, Universal Adversarial Triggers~\cite{wallace2019universal}, and Random KG Deletion.
Full attack and baseline details are in Appendix~\ref{appendix:attack-details}.

\subsection{Main Results by Attacker Access Level}
\label{subsec:rq1-main-results}

To answer RQ1, Table~\ref{tab:answer-f1-extended} reports Answer F1 for the three main datasets, three retrievers, and two generator models.
The corresponding supporting-fact Retrieval F1 results are reported in Appendix~\ref{appendix:retrieval-f1} (Table~\ref{tab:retrieval-f1-compact}).
We use these results to examine how attacker access affects the generated answer and evidence retrieved by GraphRAG under increasingly permissive threat models.

Our attacks produce larger and more consistent degradation than the existing baselines.
Even \emph{BB Query}, which uses only ranking feedback, yields lower Answer F1 than most of the six attack baselines in every evaluated dataset--retriever--generator setting.
\emph{WB Query} and \emph{Graph-aware WB} further reduce the overall average Answer F1, with \emph{Graph-aware WB} achieving the lowest average under both generators.
Several baselines substantially affect answers while changing supporting-fact retrieval only slightly; in contrast, our attacks reduce both Answer F1 and Retrieval F1.
The concurrent declines provide initial evidence that seed-ranking perturbations may affect downstream evidence selection for subsequent answer generation.

Greater access provides additional ways to disrupt the retrieval pipeline.
For example, on HotpotQA-Distractor with Contriever and LLaMA3-8B, Answer F1 falls from 52.3 in the clean setting to 37.6 under \emph{BB Query}, 26.2 under \emph{WB Query}, and 23.7 under \emph{Graph-aware WB}.
Across all 18 dataset--retriever--generator settings, \emph{WB Query} yields lower Answer F1 than \emph{BB Query}, indicating that retriever gradients strengthen the seed-ranking attack.
Adding bridge-triplet cuts also lowers supporting-fact Retrieval F1 relative to \emph{WB Query} in every setting.
The three tiers expose successive points of vulnerability: ranking feedback guides query edits, gradients sharpen them, and graph access disrupts evidence propagation through the query-specific evidence graph.

These patterns extend across the three datasets, all three retrievers, and both generator models.
The magnitude of the decline varies by setting, but ranking feedback alone is sufficient to weaken GraphRAG, while access to retriever gradients and graph structure increases the overall impact.
The observed vulnerability is therefore not tied to any single dataset, retriever, or generator model.
\begin{figure}[t]
    \centering
    \includegraphics[width=0.48\textwidth]{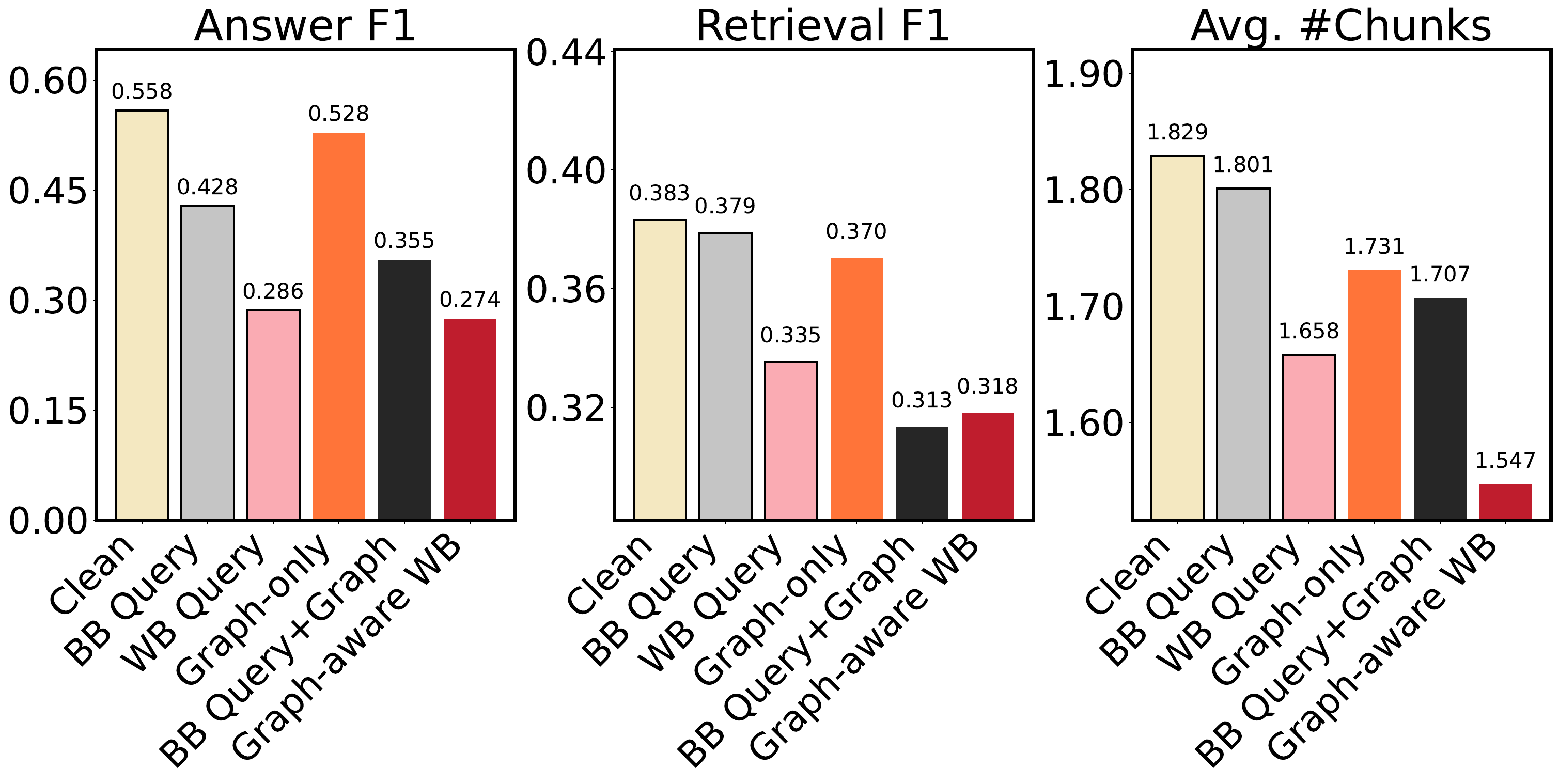}
    \caption{Component ablation on HotpotQA-Distractor. Bars show Answer F1, supporting-fact Retrieval F1, and mean retrieved chunks under clean, query-only, graph-only, and combined attacks.}
    \label{fig:ablation-bar}
\end{figure}

\subsection{Query--Graph Interaction Analysis}
\label{subsec:rq2-component-analysis}

To address RQ2, we analyze the role of graph-side perturbation in GraphRAG.
The main \method{} evaluation includes three attack settings: \emph{BB Query}, \emph{WB Query}, and \emph{Graph-aware WB}.
Here, we additionally introduce two diagnostic variants, \emph{Graph-only} and \emph{BB Query + Graph}, to isolate the effect of bridge-triplet perturbation and examine how it interacts with query-side attacks.
We conduct controlled experiments on HotpotQA-Distractor, a noisy multi-hop reasoning setting.
Figure~\ref{fig:ablation-bar} reports Answer F1, Retrieval F1, and the average number of retrieved chunks.

With BGE and LLaMA3-8B, the clean system has an Answer F1 of 0.558, a Retrieval F1 of 0.383, and 1.829 retrieved chunks on average.
Query-side attacks cause substantial degradation: \emph{WB Query} reduces Answer F1 to 0.286 and Retrieval F1 to 0.335, while \emph{BB Query} lowers Answer F1 but leaves Retrieval F1 nearly unchanged at 0.379.
\emph{Graph-only} yields an Answer F1 of 0.528 and a Retrieval F1 of 0.370.
These results establish the reference points for assessing what changes when query and graph perturbations are combined.

The combined effect on retrieval is particularly clear for \emph{BB Query}.
Adding bridge-triplet cuts reduces Retrieval F1 from 0.379 under \emph{BB Query} alone to 0.313 under \emph{BB Query + Graph}; by comparison, \emph{Graph-only} reduces it from 0.383 to 0.370.
The white-box combination also strengthens the attack: relative to \emph{WB Query}, \emph{Graph-aware WB} lowers Answer F1 from 0.286 to 0.274 and Retrieval F1 from 0.335 to 0.318.
In this setting, bridge-triplet cuts have a greater effect on evidence retrieval when paired with a perturbed query.

One plausible explanation for this observed interaction is that graph expansion can partly recover relevant evidence after a mild query shift, as suggested by the nearly unchanged Retrieval F1 under \emph{BB Query} alone.
Cutting bridge triplets may weaken these recovery paths, making the combined intervention more consequential for evidence selection.
Thus, the ablation supports a complementary role for query edits and graph cuts in disrupting GraphRAG's evidence propagation.

\subsection{Attack Behavior and Robustness}
\label{subsec:rq3-attack-behavior}

\begin{figure}[H]
    \centering
    \includegraphics[width=0.48\textwidth]{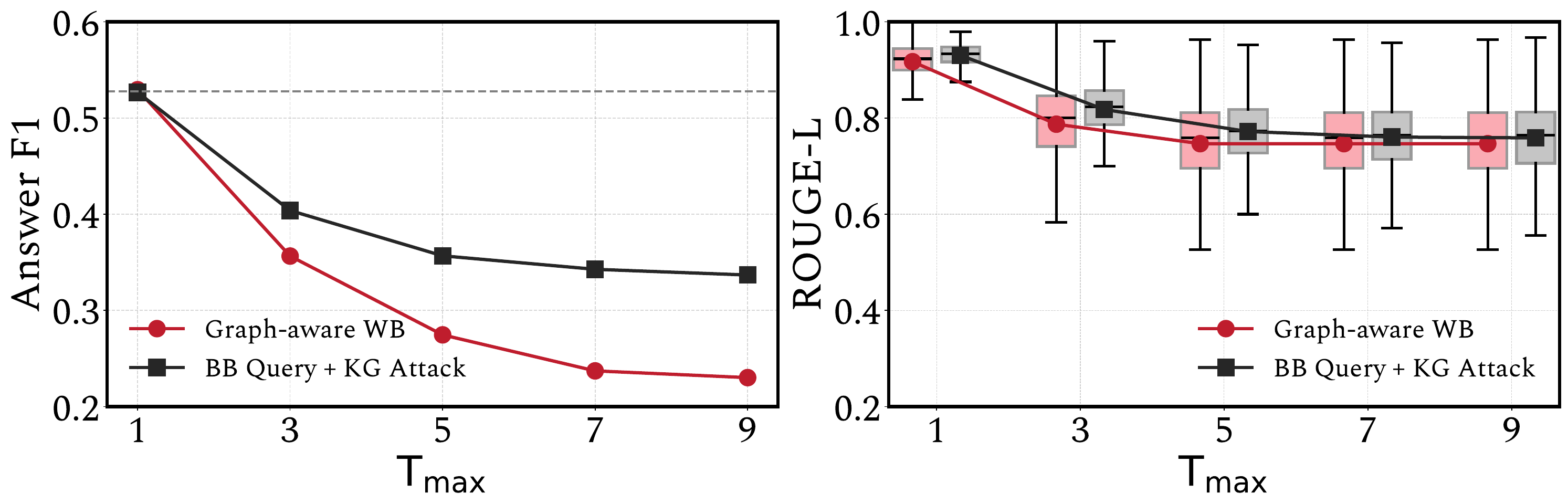}
    \caption{Answer F1 and ROUGE-L versus query substitution budget on HotpotQA-Distractor. Both variants use 5\% bridge-triplet cuts, paired with WB Query (Graph-aware WB) or BB Query (BB Query + KG Attack). Main-table results use $T_{\max}=5$.}
    \label{fig:deletion-perf}
\end{figure}

To address RQ3, we examine how \method{} behaves across substitution budgets, bridge-triplet deletion ratios, and LLM backbones.
\paragraph{Effect of Query Substitution Budget.}
Figure~\ref{fig:deletion-perf} shows that Answer F1 decreases as query substitutions increase, with gradient-guided WB updates producing a steeper decline than BB updates.
The effect levels off at larger budgets.
ROUGE-L remains relatively high, indicating that the edited queries retain lexical overlap alongside the embedding-similarity constraint.

\begin{figure}[H]
    \centering
    \includegraphics[width=0.48\textwidth]{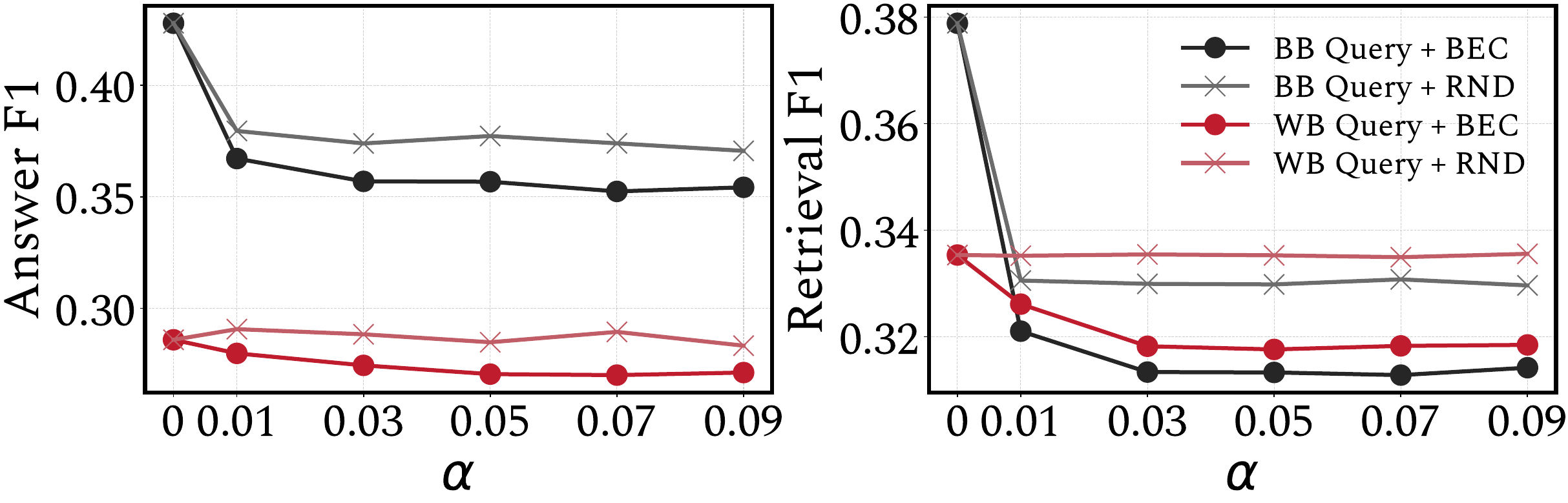}
    \caption{Answer F1 and supporting-fact Retrieval F1 versus graph deletion ratio $\alpha$ on HotpotQA-Distractor (BGE, LLaMA3-8B; $T_{\max}=5$). BEC cuts bridge triplets; RND deletes the same number randomly.}
    \label{fig:deletion-query-perf}
\end{figure}

\paragraph{Effect of Graph Deletion Ratio.}
Figure~\ref{fig:deletion-query-perf} shows that Bridge Edge Cut reduces Answer F1 and supporting-fact Retrieval F1 more than random deletion at matched budgets, before the effect levels off.
Bridge triplets form only a small fraction of each query-specific graph; together with Section~\ref{subsec:rq2-component-analysis}, this suggests that cutting these connectors weakens evidence recovery after query-side retrieval shifts.

\paragraph{Robustness Across LLM Backbones.}
Beyond the LLaMA3-8B and Qwen3-8B-Base settings, we evaluate five answer generators: Vicuna-7B, Gemma-2B, Llama-3.2-1B, Phi-3.5-mini-instruct, and GPT-J-6B.
On HotpotQA-Distractor, \emph{Graph-aware WB} causes substantial Answer F1 declines across all five models relative to their clean settings (Appendix~\ref{appendix:llm-robustness}, Figure~\ref{fig:llm-backbone}), showing that its impact extends beyond the main-table generators.

\Needspace{32\baselineskip}
\subsection{Transfer Across Graph Pipelines}
\label{subsec:architecture-transfer}

To address RQ4, we compare two graph-based evidence-selection mechanisms using the same document-extracted graph and HotpotQA-Distractor questions. PPR propagates relevance from retrieved entities, whereas the community variant selects passages through extractive Louvain reports. We evaluate the same attack conditions under both mechanisms.

\vspace{-1em}
\begin{table}[H]
\centering
\footnotesize
\setlength{\tabcolsep}{2.2pt}
\renewcommand{\arraystretch}{0.92}
\setlength{\abovecaptionskip}{4pt}
\newcommand{\transferdown}[1]{\nobreak\kern0.04em{\scriptsize\textcolor{BlueGreen}{\ensuremath{\downarrow}#1}}}
\newcommand{\transferdownbest}[1]{\nobreak\kern0.04em{\scriptsize\textcolor{BlueGreen}{\ensuremath{\downarrow}\textbf{#1}}}}
\noindent\resizebox{\columnwidth}{!}{%
\begin{tabular}{@{}lrrrr@{}}
\toprule
& \multicolumn{2}{c}{PPR} & \multicolumn{2}{c}{Community} \\
\cmidrule(lr){2-3}\cmidrule(l){4-5}
Method & Ans F1 & SP F1 & Ans F1 & SP F1 \\
\midrule
Clean                   & 50.12 & 29.62 & 54.16 & 32.51 \\
PromptInject            & 39.23\transferdown{10.89} & 28.87\transferdown{0.75} & 43.26\transferdown{10.90} & 31.93\transferdown{0.58} \\
UAT                     & 36.05\transferdown{14.07} & 29.35\transferdown{0.27} & 40.69\transferdown{13.47} & 32.43\transferdown{0.08} \\
IndirectPI              & 40.81\transferdown{9.31} & 28.60\transferdown{1.02} & 42.77\transferdown{11.39} & 31.94\transferdown{0.57} \\
PoisonedRAG             & 36.68\transferdown{13.44} & 28.14\transferdown{1.48} & 38.41\transferdown{15.75} & 31.73\transferdown{0.78} \\
HotFlip                 & 39.52\transferdown{10.60} & 27.95\transferdown{1.67} & 42.47\transferdown{11.69} & 31.69\transferdown{0.82} \\
RandKD                  & 49.96\transferdown{0.16} & 29.59\transferdown{0.03} & 54.14\transferdown{0.02} & 32.47\transferdown{0.04} \\
\midrule
BB Query                & 35.29\transferdown{14.83} & 27.15\transferdown{2.47} & 37.47\transferdown{16.69} & \secondcell{28.58}\transferdownbest{3.93} \\
WB Query                & \secondcell{27.96}\transferdownbest{22.16} & \secondcell{26.94}\transferdownbest{2.68} & \secondcell{30.73}\transferdownbest{23.43} & 29.20\transferdown{3.31} \\
Graph-aware WB          & \topcell{27.32}\transferdownbest{22.80} & \topcell{25.91}\transferdownbest{3.71} & \topcell{29.65}\transferdownbest{24.51} & \topcell{28.27}\transferdownbest{4.24} \\
\bottomrule
\end{tabular}}
\caption{Transfer F1 (\%) on HotpotQA-Distractor ($N=7{,}405$). Arrows beside scores show changes from Clean; blue shades mark the two lowest attacked scores, and bold changes mark the two largest drops in each column.}
\label{tab:architecture-transfer-main}
\end{table}
\vspace{-1em}

Table~\ref{tab:architecture-transfer-main} shows that all three access-tier attacks reduce Answer F1 and supporting-fact Retrieval F1 under both mechanisms, with larger drops than the compared baselines. Graph-aware WB produces the largest declines, whereas random graph deletion remains close to Clean. The pattern suggests that the attack effect is not specific to the evidence-selection rule used in the main experiments.

\section{Conclusion}
\label{sec:conclusion}

We examined GraphRAG's evidence path through three nested attacker access levels. Across three multi-hop QA datasets, three retrievers, and two extraction-and-generation model settings, ranking feedback alone was sufficient to degrade answers, while retriever gradients strengthened the query attack. When the adversarial query was held fixed, bridge-triplet cuts consistently reduced supporting-fact retrieval; the component ablation suggests that these cuts weaken evidence recovery after seed rankings shift. The attacks also remained effective under PPR and community-based evidence selection on the same extracted graph. These findings identify seed retrieval and graph evidence propagation as connected vulnerabilities in GraphRAG.

\section{Limitations}
\label{sec:limitations}

Our findings are bounded by the three attacker access levels and GraphRAG configurations evaluated here. They suggest two directions for improving robustness. At seed retrieval, independent retrieval signals and seed-evidence checks could reduce reliance on a single ranking before graph expansion. At graph processing, verifying triplet provenance and retaining alternative evidence paths could help detect missing bridge relations and recover supporting evidence. Evaluating these directions against adaptive attacks and across independently constructed graphs, languages, and tasks would clarify their generality.

\section{Ethical Considerations}

This work investigates adversarial vulnerabilities in GraphRAG under three levels of attacker access. We acknowledge that security research is dual-use. Our experiments use public benchmarks to study failures in seed retrieval and graph-based evidence selection; they do not target deployed products or collect user data.

The main experiments use HotpotQA, MuSiQue, and 2WikiMultihopQA and involve no human participants. Language models are used for knowledge-triplet extraction and answer generation. We reuse precomputed graphs and batch inference where possible to reduce repeated computation. The paper and appendix describe the evaluated models and datasets, attacker permissions, intervention locations, attack settings, and evaluation procedures so that the reported results can be examined in context.

During manuscript preparation, generative AI tools assisted with language editing, sentence-level rewriting, and organization. All scientific claims, experimental results, numerical values, methodological descriptions, and citations were checked and approved by the authors. The authors remain responsible for the accuracy, interpretation, and presentation of the paper.

\bibliography{custom}

\clearpage
\appendix

\section{Supplementary Results}
\label{appendix:supp-results}

\subsection{Supporting-fact Retrieval F1}
\label{appendix:retrieval-f1}
Table~\ref{tab:retrieval-f1-compact} complements the main Answer F1 table. It uses the same questions, retrievers, attack conditions, and separate LLaMA3-8B and Qwen3-8B-Base triplet-extraction and answer-generation settings.
\begin{table*}[t]
\centering
\setlength{\tabcolsep}{2.7pt}
\renewcommand{\arraystretch}{1.05}
\resizebox{\textwidth}{!}{%
\begin{tabular}{l|ccc|ccc|ccc|c}
\specialrule{1.2pt}{0pt}{0pt}
\rowcolor{headerbg}
& \multicolumn{3}{c|}{HotpotQA-Distractor} & \multicolumn{3}{c|}{MuSiQue} & \multicolumn{3}{c|}{2WikiMultihopQA} & \multicolumn{1}{c}{Overall} \\
\cline{2-4}\cline{5-7}\cline{8-10}\cline{11-11}
\rowcolor{headerbg}
\multirow{-2}{*}{\textbf{Method}} & Contriever & BGE & mxbai & Contriever & BGE & mxbai & Contriever & BGE & mxbai & Avg. \\
\specialrule{1.2pt}{0pt}{0pt}
\rowcolor{modelbg}
\multicolumn{11}{l}{\textbf{LLaMA3-8B}} \\
\rowcolor{gray!10}
Clean & 33.4 & 38.3 & 38.3 & 37.4 & 40.4 & 40.5 & 29.9 & 35.8 & 35.7 & 36.6 \\
PromptInject & 32.2\downchg{1.2} & 37.3\downchg{1.0} & 37.6\downchg{0.7} & 36.8\downchg{0.6} & 39.3\downchg{1.1} & 39.5\downchg{1.0} & 27.0\downchg{2.9} & 35.2\downchg{0.6} & 34.7\downchg{1.0} & 35.5\downchg{1.1} \\
\rowcolor{gray!10}
UAT & 32.4\downchg{1.0} & 38.1\downchg{0.2} & 38.0\downchg{0.3} & 36.4\downchg{1.0} & 39.9\downchg{0.5} & 40.0\downchg{0.5} & 29.2\downchg{0.7} & 35.9\upchg{0.1} & 35.8\upchg{0.1} & 36.2\downchg{0.4} \\
IndirectPI & 34.0\upchg{0.6} & 39.7\upchg{1.4} & 39.7\upchg{1.4} & 38.0\upchg{0.6} & 41.1\upchg{0.7} & 40.9\upchg{0.4} & 29.4\downchg{0.5} & 36.7\upchg{0.9} & 36.4\upchg{0.7} & 37.3\upchg{0.7} \\
\rowcolor{gray!10}
PoisonedRAG & 34.0\upchg{0.6} & 39.7\upchg{1.4} & 39.8\upchg{1.5} & 37.8\upchg{0.4} & 40.9\upchg{0.5} & 40.9\upchg{0.4} & 29.4\downchg{0.5} & 36.7\upchg{0.9} & 36.5\upchg{0.8} & 37.3\upchg{0.7} \\
HotFlip & 34.1\upchg{0.7} & 39.8\upchg{1.5} & 39.8\upchg{1.5} & 37.9\upchg{0.5} & 41.1\upchg{0.7} & 41.0\upchg{0.5} & 29.5\downchg{0.4} & 36.7\upchg{0.9} & 36.5\upchg{0.8} & 37.4\upchg{0.8} \\
RandKD & 33.4 & 38.4\upchg{0.1} & 38.3 & 37.4 & 40.4 & 40.5 & 29.7\downchg{0.2} & 35.6\downchg{0.2} & 35.7 & 36.6 \\
\specialrule{0.8pt}{0pt}{0pt}
\rowcolor{gray!10}
BB Query & 29.0\downchg{4.4} & 37.8\downchg{0.5} & \secondcell{33.7}\downchg{4.6} & 33.7\downchg{3.7} & \topcell{35.7}\downchg{4.7} & \topcell{35.6}\downchg{4.9} & 26.8\downchg{3.1} & 33.2\downchg{2.6} & \secondcell{32.9}\downchg{2.8} & 33.2\downchg{3.4} \\
WB Query & \secondcell{28.6}\downchg{4.8} & \secondcell{33.5}\downchg{4.8} & 34.8\downchg{3.5} & \secondcell{32.9}\downchg{4.5} & 36.3\downchg{4.1} & 37.5\downchg{3.0} & \secondcell{25.4}\downchg{4.5} & \secondcell{32.7}\downchg{3.1} & 33.1\downchg{2.6} & \secondcell{32.8}\downchg{3.8} \\
\rowcolor{gray!10}
Graph-aware WB & \topcell{26.1}\downchg{7.3} & \topcell{31.8}\downchg{6.5} & \topcell{33.3}\downchg{5.0} & \topcell{32.0}\downchg{5.4} & \secondcell{36.0}\downchg{4.4} & \secondcell{37.1}\downchg{3.4} & \topcell{22.4}\downchg{7.5} & \topcell{29.3}\downchg{6.5} & \topcell{29.8}\downchg{5.9} & \topcell{30.9}\downchg{5.7} \\
\specialrule{1.2pt}{0pt}{0pt}
\rowcolor{modelbg}
\multicolumn{11}{l}{\textbf{Qwen3-8B-Base}} \\
\rowcolor{gray!10}
Clean & 34.3 & 38.4 & 38.3 & 37.9 & 40.7 & 40.7 & 31.4 & 38.8 & 38.9 & 37.7 \\
PromptInject & 33.0\downchg{1.3} & 37.8\downchg{0.6} & 37.9\downchg{0.4} & 37.3\downchg{0.6} & 39.8\downchg{0.9} & 39.7\downchg{1.0} & 28.3\downchg{3.1} & 38.4\downchg{0.4} & 38.1\downchg{0.8} & 36.7\downchg{1.0} \\
\rowcolor{gray!10}
UAT & 33.4\downchg{0.9} & 38.3\downchg{0.1} & 38.2\downchg{0.1} & 36.8\downchg{1.1} & 40.0\downchg{0.7} & 40.2\downchg{0.5} & 30.7\downchg{0.7} & 39.1\upchg{0.3} & 38.7\downchg{0.2} & 37.3\downchg{0.4} \\
IndirectPI & 34.0\downchg{0.3} & 39.5\upchg{1.1} & 39.4\upchg{1.1} & 38.4\upchg{0.5} & 41.2\upchg{0.5} & 41.1\upchg{0.4} & 30.3\downchg{1.1} & 38.9\upchg{0.1} & 38.9 & 38.0\upchg{0.3} \\
\rowcolor{gray!10}
PoisonedRAG & 34.0\downchg{0.3} & 39.5\upchg{1.1} & 39.3\upchg{1.0} & 38.3\upchg{0.4} & 41.3\upchg{0.6} & 41.0\upchg{0.3} & 30.3\downchg{1.1} & 38.8 & 38.9 & 37.9\upchg{0.2} \\
HotFlip & 34.1\downchg{0.2} & 39.5\upchg{1.1} & 39.4\upchg{1.1} & 38.4\upchg{0.5} & 41.4\upchg{0.7} & 41.1\upchg{0.4} & 30.1\downchg{1.3} & 38.8 & 39.0\upchg{0.1} & 38.0\upchg{0.3} \\
\rowcolor{gray!10}
RandKD & 34.2\downchg{0.1} & 38.4 & 38.4\upchg{0.1} & 37.9 & 40.7 & 40.7 & 31.1\downchg{0.3} & 38.7\downchg{0.1} & 38.7\downchg{0.2} & 37.6\downchg{0.1} \\
\specialrule{0.8pt}{0pt}{0pt}
\rowcolor{gray!10}
BB Query & 29.5\downchg{4.8} & 35.2\downchg{3.2} & \secondcell{33.8}\downchg{4.5} & 34.1\downchg{3.8} & \topcell{35.9}\downchg{4.8} & \topcell{35.8}\downchg{4.9} & 27.9\downchg{3.5} & 35.5\downchg{3.3} & \secondcell{35.2}\downchg{3.7} & \secondcell{33.7}\downchg{4.0} \\
WB Query & \secondcell{29.1}\downchg{5.2} & \secondcell{34.8}\downchg{3.6} & 34.7\downchg{3.6} & \secondcell{33.3}\downchg{4.6} & 36.9\downchg{3.8} & 37.7\downchg{3.0} & \secondcell{26.4}\downchg{5.0} & \secondcell{34.6}\downchg{4.2} & 35.4\downchg{3.5} & 33.7\downchg{4.0} \\
\rowcolor{gray!10}
Graph-aware WB & \topcell{26.8}\downchg{7.5} & \topcell{33.6}\downchg{4.8} & \topcell{33.6}\downchg{4.7} & \topcell{32.1}\downchg{5.8} & \secondcell{36.5}\downchg{4.2} & \secondcell{37.2}\downchg{3.5} & \topcell{23.7}\downchg{7.7} & \topcell{31.9}\downchg{6.9} & \topcell{32.1}\downchg{6.8} & \topcell{31.9}\downchg{5.8} \\
\specialrule{1.2pt}{0pt}{0pt}
\end{tabular}}
\caption{Supporting-fact Retrieval F1 (\%) across three datasets and retrievers. Avg. is the unweighted mean of nine settings; arrows show change from \textit{Clean} within each model. Dark/light blue mark the lowest/second-lowest reported attack scores per column.}
\label{tab:retrieval-f1-compact}
\end{table*}

\subsection{TriviaQA}
\label{appendix:triviaqa}
Table~\ref{tab:triviaqa-supp} reports supplementary Answer F1 on 100 TriviaQA questions with LLaMA3-8B extraction and generation. TriviaQA lacks native sentence-level supporting-fact annotations, so we omit converted-label Retrieval F1.
\begin{table}[t]
\centering
\footnotesize
\setlength{\tabcolsep}{4.2pt}
\renewcommand{\arraystretch}{1.07}
\resizebox{\columnwidth}{!}{%
\begin{tabular}{l|ccc}
\specialrule{1.2pt}{0pt}{0pt}
\rowcolor{headerbg}
\textbf{Method} & Contriever & BGE & mxbai \\
\specialrule{1.2pt}{0pt}{0pt}
\rowcolor{gray!10}
Clean & 22.6 & 22.6 & 21.8 \\
PromptInject & 19.1\downchg{3.5} & 17.4\downchg{5.2} & 18.1\downchg{3.7} \\
\rowcolor{gray!10}
UAT & \topcell{6.2}\downchg{16.4} & \secondcell{7.6}\downchg{15.0} & \topcell{7.5}\downchg{14.3} \\
IndirectPI & 21.5\downchg{1.1} & 23.4\upchg{0.8} & 21.2\downchg{0.6} \\
\rowcolor{gray!10}
PoisonedRAG & 23.0\upchg{0.4} & 22.8\upchg{0.2} & 23.8\upchg{2.0} \\
HotFlip & 22.0\downchg{0.6} & 19.9\downchg{2.7} & 20.5\downchg{1.3} \\
RandKD & 23.3\upchg{0.7} & 22.9\upchg{0.3} & 20.9\downchg{0.9} \\
\specialrule{0.8pt}{0pt}{0pt}
\rowcolor{gray!10}
BB Query & 15.1\downchg{7.5} & 13.5\downchg{9.1} & 12.2\downchg{9.6} \\
WB Query & 9.6\downchg{13.0} & \topcell{7.2}\downchg{15.4} & 8.1\downchg{13.7} \\
\rowcolor{gray!10}
Graph-aware WB & \secondcell{9.5}\downchg{13.1} & 7.7\downchg{14.9} & \secondcell{8.0}\downchg{13.8} \\
\specialrule{1.2pt}{0pt}{0pt}
\end{tabular}}
\caption{Supplementary TriviaQA Answer F1 (\%) on 100 questions with LLaMA3-8B across three retrievers. Arrows show changes from \textit{Clean}; dark/light blue mark the lowest/second-lowest attacked scores per column.}
\label{tab:triviaqa-supp}
\end{table}

\subsection{Additional Answer Generators}
\label{appendix:llm-robustness}
\begin{figure}[t]
    \centering
    \includegraphics[width=0.92\linewidth]{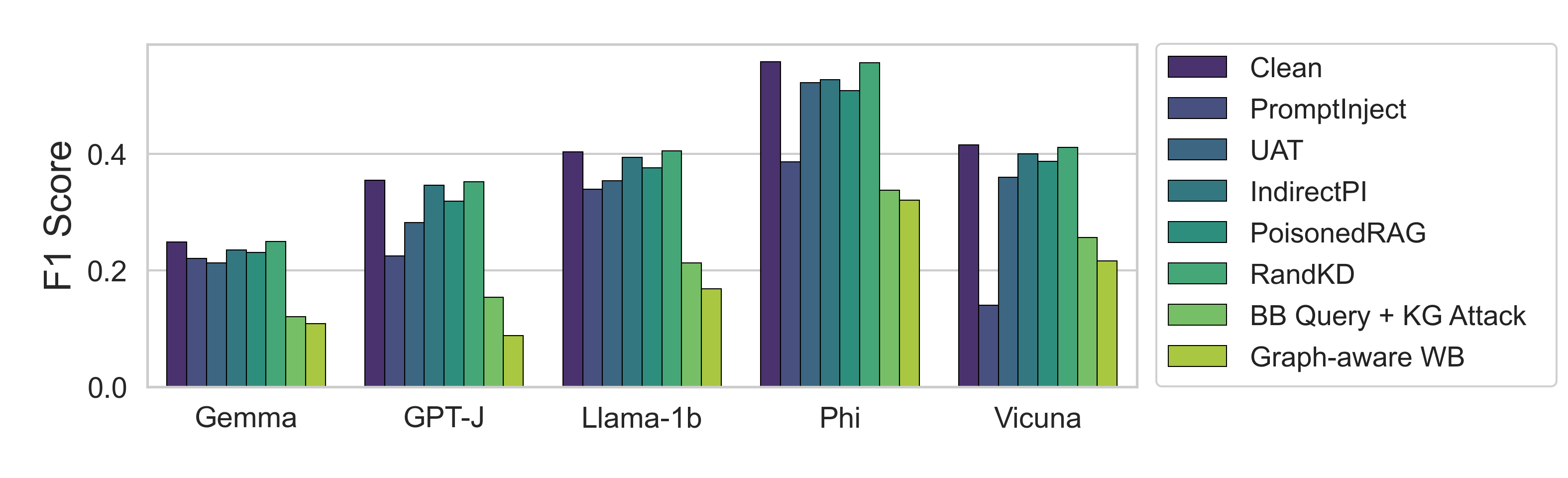}
    \caption{Answer F1 on HotpotQA-Distractor with five additional readers under the LLaMA3-8B graph-extraction setting.}
    \label{fig:llm-backbone}
\end{figure}

Figure~\ref{fig:llm-backbone} reports Answer F1 on HotpotQA-Distractor with five additional readers, holding the earlier LLaMA3-8B graph-extraction setting fixed. These are reader substitutions, whereas the Qwen3-8B-Base main-table setting changes both extraction and generation.

\section{Experimental Protocol}
\label{appendix:setup}

\paragraph{Data and graph construction.}
We follow the KG2RAG preprocessing protocol~\cite{zhu2025knowledge} for HotpotQA-Distractor (7,405 questions), MuSiQue (2,417), and TriviaQA (100). The 2WikiMultihopQA evaluation uses a fixed 1,000-question development subset converted to the same paragraph-and-sentence schema. Across datasets, triplets $(h,r,t)$ are extracted from document text and linked to source chunks; gold evidence and answers are excluded from graph construction.

\paragraph{Retrieval and models.}
KG2RAG retrieves $K=10$ seed chunks, expands their linked triplets, and reranks candidate chunks with \texttt{bge-reranker-large}; the final context has at most 10 chunks. At this fixed $K$, the exponential importance weights in Eq.~\ref{eq:dcg-weight} do not overflow; a larger-$K$ implementation can evaluate the same normalized weights with rescaling or in log space. We evaluate Contriever, BGE-large-en-v1.5, and mxbai-embed-large. The two main-table model settings use LLaMA3-8B or Qwen3-8B-Base for both triplet extraction and answer generation. Other analyses use LLaMA3-8B unless stated otherwise. Appendix~\ref{appendix:prompts} reproduces the prompts.

\paragraph{Attack settings and baselines.}
\label{appendix:attack-details}
The main query-attack budget is $T_{\max}=5$ iterations. Accepted substitutions must reduce the ranking loss and satisfy cosine similarity $\geq0.8$ under all-MiniLM-L6-v2. BB Query uses observed rankings and WordNet candidate substitutions; WB Query uses retriever gradients over a top-$(K+M)$ candidate pool with $M=50$. Graph-aware WB deletes 5\% of triplets in the query-specific evidence graph, prioritizing bridges; Random KG Deletion uses the same deletion budget. HotFlip is a \emph{corpus-poisoning} baseline: it optimizes one injected passage per question with at most 20 token-replacement steps and 32 candidates per step, capped at 96 tokens. It does not edit the query. The remaining baseline implementations follow their cited attacks.

\paragraph{Evaluation.}
\label{appendix:eval-protocol}
Answer F1 uses normalized token overlap with the gold answer; \textit{yes}, \textit{no}, and \textit{noanswer} require exact matching. Supporting-fact F1 uses exact matching of (title, sentence-index) pairs. We report means of question-level F1 values, so F1 need not equal the harmonic mean of separately averaged Precision and Recall. For paired uncertainty analyses, bootstrap resampling uses questions as the unit.

\section{Prompt Templates}
\label{appendix:prompts}
The same templates are used for LLaMA3-8B and Qwen3-8B-Base; the text chunk, retrieved context, and question replace the placeholders below.

\subsection{Triplet Extraction Prompt}
\begin{prompt}
Extract triplets informative from the text following the examples. Make sure the triplet texts are only directly from the given text! Complete directly and strictly following the instructions without any additional words, line break nor space!
--------------------
Text: Scott Derrickson (born July 16, 1966) is an American director, screenwriter and producer.
Triplets:<Scott Derrickson##born in##1966>$$<Scott Derrickson##nationality##America>$$<Scott Derrickson##occupation##director>$$<Scott Derrickson##occupation##screenwriter>$$<Scott Derrickson##occupation##producer>$$
--------------------
Text: A Kiss for Corliss is a 1949 American comedy film directed by Richard Wallace and written by Howard Dimsdale. It stars Shirley Temple in her final starring role as well as her final film appearance. Shirley Temple was named United States ambassador to Ghana and to Czechoslovakia and also served as Chief of Protocol of the United States.
Triplets:<A Kiss for Corliss##cast member##Shirley Temple>$$<Shirley Temple##served as##Chief of Protocol>$$
--------------------
Text: {ctx_text}
Triplets:
\end{prompt}

\subsection{Answer Generation Prompt}
\begin{prompt}
Context information:
{context_str}

Q: Were Scott Derrickson and Ed Wood of the same nationality?
A: Yes
Q: Who was born earlier, Emma Bull or Virginia Woolf?
A: Adeline Virginia Woolf
Q: The arena where the Lewiston Maineiacs played their home games can seat how many people?
A: 3,677 seated
Q: What government position was held by the woman who portrayed Corliss Archer in the film Kiss and Tell?
A: Chief of Protocol
Q: Which U.S. state has the capital city named Helena?
A: Montana
Q: {query_str}
A:
\end{prompt}

\end{document}